\documentclass[11pt,a4paper]{article}

\usepackage[usenames,dvipsnames,svgnames,table]{xcolor}  
\usepackage{todonotes} 

\usepackage[hyperref]{naaclhlt2019}
\usepackage{times}
\usepackage{adjustbox}
\usepackage{latexsym}
\usepackage{gb4e}
\usepackage{xfrac}
\noautomath
\usepackage{CJKutf8}
\usepackage{inconsolata}
\usepackage{amsmath}
\usepackage{url}
\usepackage{booktabs}
\usepackage{microtype}
\usepackage{pinyin}
\usepackage[safe]{tipa}
\usepackage{graphicx}
\usepackage[printwatermark]{xwatermark}
\graphicspath{ {/} }

\aclfinalcopy 
\def\aclpaperid{379} 

\usepackage{cleveref}
\crefname{section}{\S}{\S\S}
\Crefname{section}{\S}{\S\S}
\crefname{table}{Table}{}
\crefname{figure}{Figure}{}
\crefname{algorithm}{Algorithm}{}
\crefname{equation}{eq.}{}
\crefname{appendix}{App.}{}
\crefformat{section}{\S#2#1#3}  
\usepackage{url}

\newcommand{\defn}[1]{\textbf{#1}}
\newcommand{\word}[1]{\textit{#1}}

\title{On the Idiosyncrasies of the Mandarin Chinese Classifier System}

\author{Shijia Liu$^{\textrm{\normalfont\textschwa}}$ \quad Hongyuan Mei$^{\textrm{\normalfont \textschwa}}$ \quad Adina Williams$^{\textrm{\normalfont \textipa{P}}}$ \quad Ryan Cotterell$^{\textrm{\normalfont \textschwa, \textipa{H}}}$ \\
${}^{\textrm{\textschwa}}$Department of Computer Science, Johns Hopkins University \\
${}^{\textrm{\textipa{P}}}$Facebook Artificial Intelligence Research \\
${}^{\textrm{\textipa{H}}}$Department of Computer Science and Technology, University of Cambridge  \\
{\tt \{\href{mailto:sliu126@jhu.edu}{sliu126},\href{mailto:hmei@jhu.edu}{hmei}\}@jhu.edu} \quad  {\tt \href{mailto:adinawilliams@fb.com}{adinawilliams@fb.com}} \quad {\tt \href{mailto:rdc42@cam.ac.uk}{rdc42@cam.ac.uk}}
}

\date{}

\begin{document}
\maketitle
\begin{abstract}

While idiosyncrasies of the Chinese classifier system have been a richly studied topic among linguists \citep{adams1973,erbaugh1986,lakoff1986}, not much work has been done to quantify them with statistical methods. In this paper, we introduce an information-theoretic approach to measuring idiosyncrasy; we examine how much the uncertainty in Mandarin Chinese classifiers can be reduced by knowing semantic information about the nouns that the classifiers modify. Using the empirical distribution of classifiers from the parsed Chinese Gigaword corpus \cite{Graff:05}
, we compute the mutual information (in bits) between the distribution over classifiers and distributions over other linguistic quantities. We investigate whether semantic classes of nouns and adjectives differ in how much they reduce uncertainty in classifier choice, and find that it is not fully idiosyncratic; while there are no obvious trends for the majority of semantic classes, shape nouns reduce uncertainty in classifier choice the most.
\end{abstract}

\section{Introduction}\label{sec:introduction}
Many of the world's languages make use of \defn{numeral classifiers} \cite{Aikhenvald-2000}. While theoretical debate still rages on the function of numeral classifiers \citep{krifka1995,ahrens1996,cheng1998,chierchia1998,li2000,nisbett2004,bale2014}, it is generally accepted that they need to be present for nouns to be modified by numerals, quantifiers, demonstratives, or other qualifiers \citep[104]{li1981}. In Mandarin Chinese, for instance, the phrase \word{one person} translates as \begin{CJK}{UTF8}{gbsn}一个人\end{CJK} (\begin{CJK}{UTF8}{gbsn}\textit{\yi1 \ge4 \ren2}\end{CJK}); the classifier \begin{CJK}{UTF8}{gbsn}个\end{CJK} (\textit{g{\`e}}) has no clear translation in English, yet, nevertheless, it is necessary to place it between the numeral 
\begin{CJK}{UTF8}{gbsn}一\end{CJK} (\begin{CJK}{UTF8}{gbsn}\textit{\yi1}\end{CJK}) and the word for person \begin{CJK}{UTF8}{gbsn}人\end{CJK} (\textit{r{\'e}n}). \looseness=-1

\begin{table}
\centering
\begin{adjustbox}{width=\columnwidth}
\begin{tabular}{lll}
\toprule
{\bf Classifier} & \textit{\bf \begin{CJK}{UTF8}{gbsn}\Pin1\yin1\end{CJK}} & {\bf Semantic Class } \\
\midrule
\begin{CJK}{UTF8}{gbsn}个\end{CJK} & \textit{\begin{CJK}{UTF8}{gbsn}\ge4\end{CJK}} & objects, general-purpose \\
\begin{CJK}{UTF8}{gbsn}件\end{CJK} & \textit{\begin{CJK}{UTF8}{gbsn}\jian4\end{CJK}} & matters \\
\begin{CJK}{UTF8}{gbsn}头\end{CJK} & \textit{\begin{CJK}{UTF8}{gbsn}\tou2\end{CJK}} & domesticated animals \\
\begin{CJK}{UTF8}{gbsn}只\end{CJK} & \textit{\begin{CJK}{UTF8}{gbsn}\zhi1\end{CJK}} & general animals \\
\begin{CJK}{UTF8}{gbsn}张\end{CJK} & \textit{\begin{CJK}{UTF8}{gbsn}\zhang1\end{CJK}} & flat objects \\
\begin{CJK}{UTF8}{gbsn}条\end{CJK} & \textit{\begin{CJK}{UTF8}{gbsn}\tiao2\end{CJK}} & long, narrow objects \\
\begin{CJK}{UTF8}{gbsn}项\end{CJK} & \textit{\begin{CJK}{UTF8}{gbsn}\xiang4\end{CJK}} & items, projects \\
\begin{CJK}{UTF8}{gbsn}道\end{CJK} & \textit{\begin{CJK}{UTF8}{gbsn}\dao4\end{CJK}} & orders, roads, projections \\
\begin{CJK}{UTF8}{gbsn}匹\end{CJK} & \textit{\begin{CJK}{UTF8}{gbsn}\pi3\end{CJK}} & horses, cloth \\
\begin{CJK}{UTF8}{gbsn}顿\end{CJK} & \textit{\begin{CJK}{UTF8}{gbsn}\dun4\end{CJK}} & meals \\
\bottomrule
\end{tabular}
\end{adjustbox}
\caption{Examples of Mandarin classifiers. Classifiers' simplified Mandarin Chinese characters (1\textsuperscript{st} column), \textit{\begin{CJK}{UTF8}{gbsn}\pin1\end{CJK}\begin{CJK}{UTF8}{gbsn}\yin1\end{CJK}} pronunciations (2\textsuperscript{nd} column), and commonly modified noun types (3\textsuperscript{rd} column) are listed.}
\label{tbl:classifier}
\end{table} 

There are hundreds of numeral classifiers in the Mandarin lexicon \citep[\cref{tbl:classifier} gives some canonical examples]{po2015chinese}, and classifier choices are often argued to be  based on inherent, possibly universal, semantic properties associated with the noun, such as shape \cite{Kuo:09,Zhang:16}. Indeed, in a summary article,
\newcite{tai1994chinese} writes: ``Chinese classifier systems are cognitively based, rather than arbitrary systems of classification.'' 
If classifier choice were solely based on conceptual features of a given noun, then we might expect it to be nearly determinate---like gender-marking on nouns in German, Slavic, or Romance languages \citep[400]{erbaugh1986}---and perhaps even fixed for all of a given noun's synonyms.\looseness=-1

However, selecting which classifiers go with nouns in practice is an idiosyncratic process, often with several grammatical options (see \cref{tbl: empirical distribution} for two examples). Moreover, knowing what a noun means does not always mean you can guess the classifier. For example, most nouns referring to animals, such as \begin{CJK}{UTF8}{gbsn}驴\end{CJK} (\begin{CJK}{UTF8}{gbsn}\textit{\lv2}\end{CJK}, \textit{donkey}) or \begin{CJK}{UTF8}{gbsn}羊\end{CJK} (\begin{CJK}{UTF8}{gbsn}\textit{\yang2}\end{CJK}, \textit{goat}), use the classifier \begin{CJK}{UTF8}{gbsn}只\end{CJK} (\begin{CJK}{UTF8}{gbsn}\textit{\zhi1}\end{CJK}). However, horses cannot use \begin{CJK}{UTF8}{gbsn}只\end{CJK} (\begin{CJK}{UTF8}{gbsn}\textit{\zhi1}\end{CJK}), despite being semantically similar to goats and donkeys, and instead must appear with \begin{CJK}{UTF8}{gbsn}匹\end{CJK} (\begin{CJK}{UTF8}{gbsn}\textit{\pi3}\end{CJK}). Conversely, knowing which particular subset of noun meaning is reflected in the classifier also does not mean that you can use that classifier with any noun that seems to have the right semantics. For example, the classifier \begin{CJK}{UTF8}{gbsn}条\end{CJK} (\textit{ti{\'a}o}) can be used with nouns referring to long and narrow objects, like rivers, snakes, fish, pants, and certain breeds of dogs---but never cats, regardless of how long and narrow they might be! In general, classifiers carve up the semantic space in a very idiosyncratic manner that is neither fully arbitrary, nor fully predictable from semantic features.  \looseness=-1

 \begin{table}
 \centering
 \begin{adjustbox}{width=\columnwidth}
 \begin{tabular}{lll}
 \toprule
 {\bf Classifier} & {$p(C \mid N = \textbf{\begin{CJK}{UTF8}{gbsn}人士\end{CJK}})$} & {$p(C \mid N = \textbf{\begin{CJK}{UTF8}{gbsn}工程\end{CJK}})$} \\
 \midrule
 \begin{CJK}{UTF8}{gbsn}位\end{CJK} (\begin{CJK}{UTF8}{gbsn}\textit{\wei4}\end{CJK}) & \bf 0.4838 & 0.0058 \\
 \begin{CJK}{UTF8}{gbsn}名\end{CJK} (\begin{CJK}{UTF8}{gbsn}\textit{\ming2}\end{CJK}) & \bf 0.3586 & 0.0088 \\
 \begin{CJK}{UTF8}{gbsn}个\end{CJK} (\begin{CJK}{UTF8}{gbsn}\textit{\ge4}\end{CJK}) & 0.0205 & 0.0486 \\
 \begin{CJK}{UTF8}{gbsn}批\end{CJK} (\begin{CJK}{UTF8}{gbsn}\textit{\pi1}\end{CJK}) & 0.0128 & 0.0060 \\
 \begin{CJK}{UTF8}{gbsn}项\end{CJK} (\begin{CJK}{UTF8}{gbsn}\textit{\xiang4}\end{CJK}) & 0.0063 & \bf 0.4077 \\
 \begin{CJK}{UTF8}{gbsn}期\end{CJK} (\begin{CJK}{UTF8}{gbsn}\textit{\qi1}\end{CJK}) & 0.0002 & \bf 0.2570 \\
  Everything else & 0.1178 & 0.2661 \\
 \bottomrule
 \end{tabular}
 \end{adjustbox}
 \caption{Empirical distribution of selected classifiers over two nouns: \begin{CJK}{UTF8}{gbsn}人士 (\textit{\ren2 \shi4, people})\end{CJK} and \begin{CJK}{UTF8}{gbsn}工程 (\textit{\gong1 \cheng2, project})\end{CJK}. Most of the probability mass is allocated to only a few classifiers for both nouns (bolded). }
 \label{tbl: empirical distribution}
 \end{table}  

Given this, we can ask: precisely \emph{how} idiosyncratic is the Mandarin Chinese classifier system? For a given noun, how predictable is the set of classifiers that can be grammatically employed? For instance, had we not known that the Mandarin word for horse \begin{CJK}{UTF8}{gbsn}马\end{CJK} (\begin{CJK}{UTF8}{gbsn}\textit{\ma3}\end{CJK}) predominantly takes the classifier
\begin{CJK}{UTF8}{gbsn}匹\end{CJK} (\begin{CJK}{UTF8}{gbsn}\textit{\pi3}\end{CJK}), how likely would we have been to guess it over 
the much more common animal classifier \begin{CJK}{UTF8}{gbsn}只\end{CJK} (\begin{CJK}{UTF8}{gbsn}\textit{\zhi1}\end{CJK})? Is it more important to know that a noun is \begin{CJK}{UTF8}{gbsn}马\end{CJK} (\begin{CJK}{UTF8}{gbsn}\textit{\ma3}\end{CJK}) or simply that the noun is an animal noun? We address these questions by computing how much the uncertainty in the distribution over classifiers can be reduced by knowing information about nouns and noun semantics. We quantify this notion of classifier idiosyncrasy by calculating the mutual information between classifiers and nouns, and also between classifiers and several sets that are relevant to noun meaning (i.e.,\ categories of noun senses, sets of noun synonyms, adjectives, and categories of adjective senses). 
Our results yield concrete, quantitative measures of idiosyncrasy in bits, that can supplement existing hand-annotated, intuition-based approaches that organize Mandarin classifiers into an ontology.\looseness=-1

Why investigate the idiosyncrasy of the Mandarin Chinese classifier system? How idiosyncratic or predictable natural language is has captivated researchers since \newcite{shannon1951prediction} originally proposed the question in the context of printed English text. Indeed, looking at predictability directly relates to the complexity of language---a fundamental question in linguistics \cite{newmeyer2014measuring,dingemanse2015arbitrariness}---which has also been claimed to have consequences for learnability and processing. For example, how hard it is for a learner to master irregularity, say, in the English past tense \citep{rumelhart1987,pinker1994,pinker2002, kirov2018} might be affected by predictability, and highly predictable noun-adjacent words, such as gender affixes in German and pre-nominal adjectives in English, are also shown to confer online processing advantages \citep{dye2016,dye2017,dye2018}. Within the Chinese classifier system itself, the very common, general-purpose classifier \begin{CJK}{UTF8}{gbsn}个\end{CJK} (\textit{\begin{CJK}{UTF8}{gbsn}\ge4\end{CJK}}) is acquired by children earlier than rarer, more semantically rich ones \citep{hu1993}. General classifiers are also found to occur more often in corpora with nouns that are less predictable in context (i.e., nouns with high \textit{surprisal}; \citealt{hale2001}) \citep{zhan2018}, providing initial evidence that predictability likely plays a role in classifier-noun pairing more generally. Furthermore, providing classifiers improves participants' recall of nouns in laboratory experiments (\citealp{zhang1998, gao2009}; see \citealp{huang2014})---but, it is not known whether classifiers do so by modulating noun predictability.\looseness=-1

\section{Quantifying Classifier Idiosyncrasy}

We take an information-theoretic approach to statistically quantify the idiosyncrasy of the Mandarin Chinese classifier system, and measure the uncertainty (entropy) reduction---or \defn{mutual information} \cite[MI;][]{cover2012elements}---between classifiers and other linguistic quantities, like nouns or adjectives.
Intuitively, MI lets us directly measure classifier ``idiosyncrasy'', because it tells us how much information (in bits) about a classifier we can get by observing another linguistic quantity. 
If classifiers were completely independent from other quantities, knowing them would give us no information about classifier choice.

\paragraph{Notation.}
Let $C$ be a classifier-valued random variable with range $\mathcal{C}$,
the set of Mandarin Chinese classifiers.
Let $X$ be a second random variable, which models a second linguistic quantity, with range $\mathcal{X}$. Mutual information (MI) is defined as
    \begin{subequations}
    \begin{align}
    I(C; X) &\equiv H(C) - H(C \mid X) \\
    		&= \sum_{c \in \mathcal{C}} \sum_{x \in \mathcal{X}} p(c, x) \log \frac{p(c,x)}{p(c)p(x)}
    \end{align}
    \end{subequations}
Let $\mathcal{N}$ and $\mathcal{A}$ denote the sets of nouns and adjectives, respectively, and let $N$ and $A$ be noun- and adjective-valued random variables, respectively. Let $\mathcal{N}_i$ and $\mathcal{A}_i$ denote the sets of nouns and adjectives in the $i$\textsuperscript{th} SemCor supersense category for nouns \citep{qvec:enmlp:15} and adjectives \citep{Tsvetkov:14}, respectively, with their random variables being $N_i$ and $A_i$, respectively.
Let $\mathcal{S}$ be the set of all English WordNet \cite{miller1998wordnet} senses of nouns, with $S$ being the WordNet sense-valued random variable. Given the formula above and any choice for $X \in \{N, A, N_i, A_i, S\}$, we can calculate the mutual information between classifiers and other relevant linguistic quantities.

\subsection{Classifiers and Nouns}
 Mutual information between classifiers and nouns $(I(C;N))$ shows how much uncertainty (i.e.,\ entropy) in classifiers can be reduced once we know the noun, and vice versa. Because only a few classifiers are suitable to modify a given noun (again, see \cref{tbl: empirical distribution}) and the entropy of classifiers for a given noun is predicted to be close to zero, MI between classifiers and nouns is expected to be high. 
     
\subsection{Classifiers and Adjectives}
The distribution over adjectives that modify a noun in a large corpus
gives us a language-internal peek into a word's lexical semantics. 
Moreover, adjectives have been found to increase the predictability of nouns \citep{dye2018}, so we ask whether they might affect classifier predictability too. We compute MI between classifiers and adjectives ($I(C;A)$) that modify the same nouns to investigate their relationship. If both adjectives and classifiers track comparable portions of the noun's semantics, we expect $I(C;A)$ to be significantly greater than zero, which implies mutual dependence between classifier $C$ and adjective $A$. 

 \subsection{Classifiers and Noun Supersenses}
To uncover which nouns are able to reduce the uncertainty in classifiers more, we divide them into 26 SemCor supersense categories ~\citep{qvec:enmlp:15}, and then compute $I(C;N_i)$ for $i \in \{1,2, \ldots, 26\}$ for each supersense category. The supersense categories (e.g.,\ {\em animal, plant, person, artifact}, etc.) provide a semantic classification system for English nouns. Since there are no known supersense categories for Mandarin, we need to translate Chinese nouns into English to perform our analysis. We use SemCor supersense categories instead of WordNet hypernyms because different ``basic levels'' for each noun make it difficult to determine the ``correct'' category for each noun. 

\subsection{Classifiers and Adjective Supersenses}
We translated and divided the adjectives into 12 \defn{supersense categories} \citep{Tsvetkov:14}, and compute mutual information $I(C;A_i)$ for categories $i \in \{1,2, \ldots, 12\}$ to determine which categories have more mutual dependence on classifiers. Adjective supersenses are defined as categories describing certain properties of nouns. For example, adjectives in the \textsc{mind} category describe intelligence and awareness, while those in the \textsc{perception} category focus on, e.g.,\ color, brightness, and taste.
Examining the distribution over adjectives is a language-specific measure of \textit{noun} meaning, albeit an imperfect one, because only certain adjectives modify any given noun. 
     
\subsection{Classifiers and Noun Synsets}
We also compute the mutual information $I(C;S)$ between classifiers and nouns' WordNet ~\cite{miller1998wordnet} synonym sets (\defn{synsets}), assuming that each synset is independent. For nouns with multiple synsets, we assume that all synsets are equally probable for simplicity. If classifiers are fully semantically determined, then knowing a noun's synsets should enable one to know the appropriate classifier(s), resulting in high MI. If classifiers are largely idiosyncratic, then noun synsets should have lower MI with classifiers. We do not use WordNet to attempt to capture word polysemy here. 

\begin{table*}[t]
    \centering
    \begin{tabular}{lllllll}
    \toprule
    {$H(C)$} & {$H(C \mid N)$} & {$I(C; N)$} & {$H(C \mid S)$} & {$I(C; S)$} & {$H(C \mid A)$} & {$I(C; A)$} \\
    \midrule
    5.61 & 0.66 & 4.95 & 4.14 & 1.47 & 3.53 & 2.08 \\
    \bottomrule
    \end{tabular}
    \caption{Mutual information between classifiers and nouns $I(C;N)$, noun senses $I(C;S)$, and adjectives $I(C;A)$, is compared to their entropies.}
    \label{tbl: overall_mi}
\end{table*}

\section{Data and Experiments}

 \paragraph{Data Provenance.}
We apply an existing neural Mandarin word segmenter \cite{Cai:17} to the Chinese Gigaword corpus \cite{Graff:05}, and then feed the segmented corpus to a neural dependency parser,
 using Google's pretrained Parsey Universal model on Mandarin.\footnote{\url{https://github.com/tensorflow/models/blob/master/research/syntaxnet/g3doc/universal.md}} The model is trained on Universal Dependencies datasets v1.3.\footnote{\url{https://universaldependencies.org/}} 
We extract classifier-noun pairs and adjective-classifier-noun triples from sentences, where the adjective and the classifier modify the same noun---this is easily determined from the parse. We also record the tuple counts, and use them to compute an empirical distribution over classifiers that modify nouns, and noun-adjective pairs, respectively.

 \paragraph{Data Preprocessing.}
Since no annotated supersense list exists for Mandarin, we first use CC-CEDICT\footnote{\url{www.mdbg.net/chinese/dictionary}} as a Mandarin Chinese-to-English dictionary to translate nouns and adjectives into English. Acknowledging that translating might introduce noise, we subsequently categorize our words into different senses using the SemCor supersense data for nouns \citep{miller1993,qvec:enmlp:15}, and adjectives \citep{Tsvetkov:14}. We calculate the mutual information under each noun, and adjective supersense using plug-in estimation. 
    
\paragraph{Modeling Assumptions.}
As this contribution is the first to investigate classifier predictability, we make several simplifying assumptions. Extracting distributions over classifiers from a large corpus, as we do, \emph{ignores} sentential context, which means we ignore the fact that some nouns (i.e.,\ relational nouns, like \begin{CJK}{UTF8}{gbsn}\textit{\ma1}\end{CJK}\begin{CJK}{UTF8}{gbsn}\textit{\ma1}\end{CJK}, \textit{Mom}) are more likely to be found in verb frames or other constructions where classifiers are not needed.
We also ignore singular--plural, which might affect classifier choice, and the mass--count distinction \citep{cheng1998,bale2014}, to the extent that it is not encoded in noun supersense categories (e.g.,\ \emph{substance} includes mass nouns).
We also assume that every classifier--noun or classifier--adjective pairing we extract is equally acceptable to native speakers. However, it's possible that native speakers differ in either their knowledge of classifier-noun distributions or confidence in particular combinations. Whether and how such human knowledge interacts with our calculations would be an interesting future avenue.\looseness=-1

\section{Results and Analysis}\label{sec:results}

\subsection{Classifiers and Nouns, Noun Synsets, and Adjectives} \looseness=-1 \cref{tbl: overall_mi} shows MI between classifiers and other linguistic quantities. As we can see, {$I(C; N) > I(C; A) > I(C; S)$}. As expected, knowing the noun greatly reduces classifier uncertainty; the noun and classifier have high MI ($4.95$ bits). Classifier MI with noun synsets ($1.47$ bits) is not comparable to that with nouns ($4.95$ bits), suggesting that knowing a synset does not greatly reduce classifier uncertainty, leaving $>$$\sfrac{3}{5}$ of the entropy unaccounted for. 
We also see that adjectives ($2.08$ bits) reduce the uncertainty in classifiers more than noun synsets ($1.47$ bits), but less than nouns ($4.95$ bits). 

\subsection{Classifiers and Noun Supersenses}  
\begin{figure}
    	\centering
        \includegraphics[width=\columnwidth]{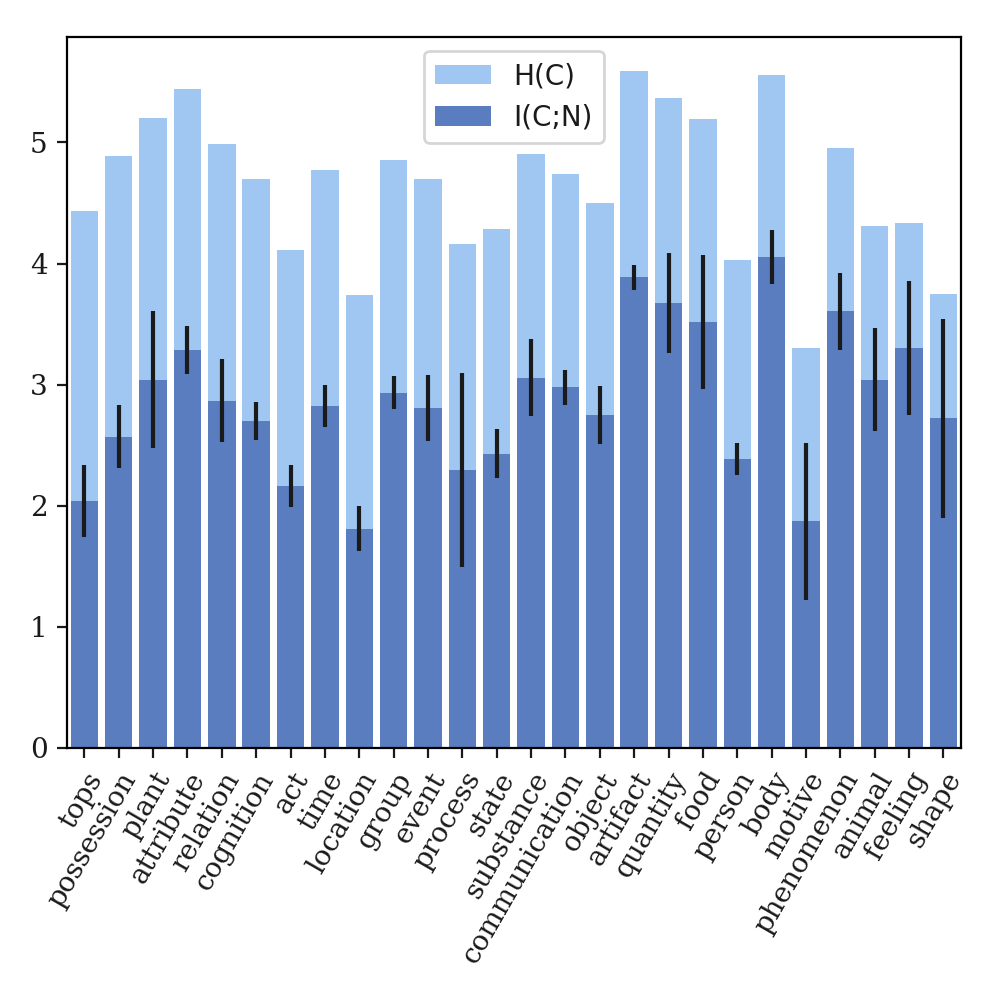}
        \caption{Mutual information between classifiers and nouns (dark blue), and classifier entropy (light \& dark blue) plotted with $H(C\mid N) = H(C) - I(C; N)$ (light blue) decreasing from left. Error bars denote  bootstrapped 95\% confidence interval of $I(C; N)$.}
    	\label{fig: mi_class_nouns}
    \end{figure}

Noun supersense results are in \cref{fig: mi_class_nouns}. Natural categories are helpful, but are far from completely predictive of the classifier distribution: knowing that a noun is a \word{plant} helps, but cannot account for about $\sfrac{1}{3}$ of the original entropy for the distribution over classifiers, and knowing that a noun is a \word{location} leaves $>$$\sfrac{1}{2}$ unexplained. The three supersenses with highest $I(C; N)$ are {\em body}, {\em artifact}, and {\em shape}. Of particular interest is the {\em shape} category. Knowing that a noun refers to a shape (e.g., \begin{CJK}{UTF8}{gbsn}角度\end{CJK}; \begin{CJK}{UTF8}{gbsn}\textit{\jiao3}\end{CJK} \begin{CJK}{UTF8}{gbsn}\textit{\du4}\end{CJK}, \word{angle}), makes the choice of classifier 
relatively predictable. This result sheds new light on psychological findings that Mandarin speakers are more likely to classify words as similar based on shape than English speakers \citep{Kuo:09}, by uncovering a possible role for information structure in shape-based choice. It also accords with \citet{chien2003} and \citet[216]{li2010} that show that children as young as three know that classifiers often delineate categories of objects with similar shapes.\looseness=-1  

     \subsection{Classifiers and Adjective Supersenses} 
\begin{figure}
    	\centering
        \includegraphics[width=\columnwidth]{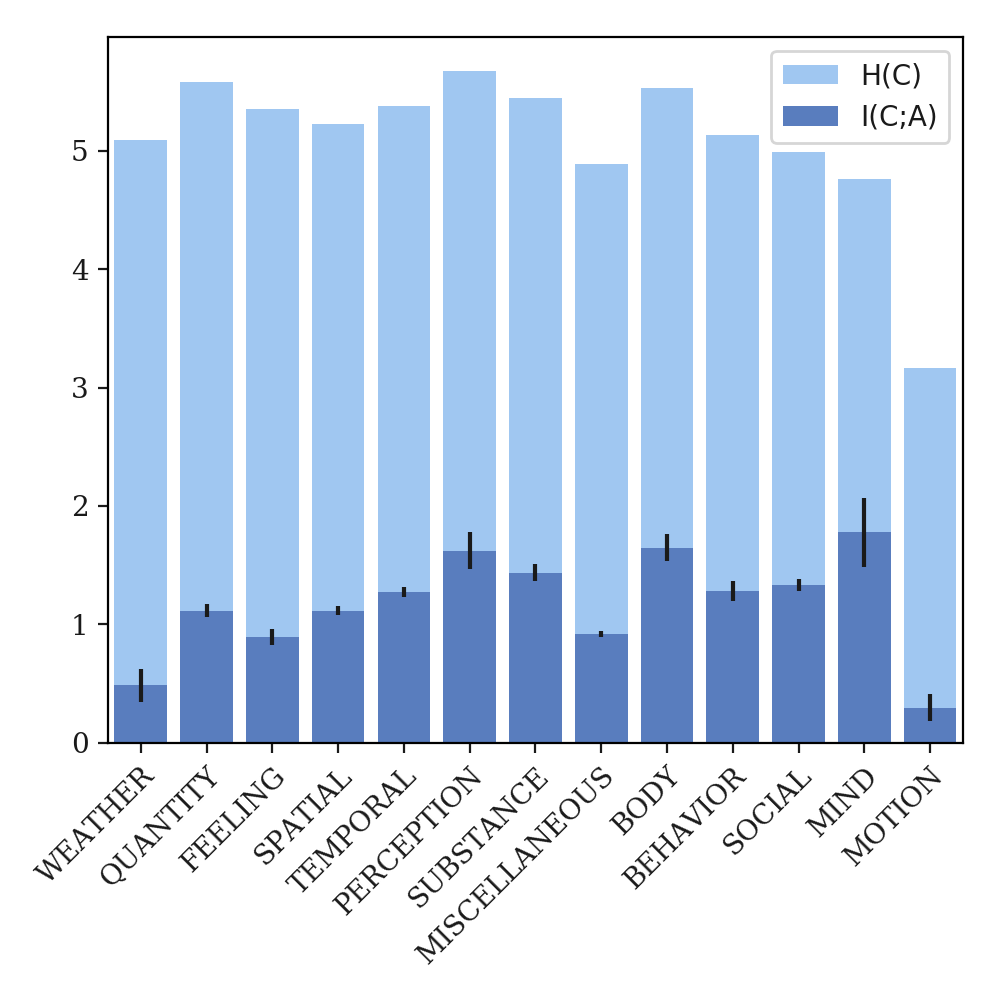}
        \caption{Mutual information between classifiers and adjectives (dark blue), and classifier entropy (light \& dark blue) plotted with $H(C\mid A) = H(C) - I(C; A)$ (light blue) decreasing from left. Error bars denote bootstrapped 95\% confidence interval of $I(C; A)$.}
    	\label{fig: mi_class_adjectives}
    \end{figure}

Adjective supersense results are in \cref{fig: mi_class_adjectives}. Interestingly, the top three senses that have the highest mutual information between their adjectives and classifiers---\textsc{mind}, \textsc{body} (constitution, appearance) and \textsc{perception}---are all involved with people's subjective views. With respect to \newcite{Kuo:09}, adjectives from the \textsc{spatial} sense pick out shape nouns in our results. Although it does not make it into the top three, MI for the \textsc{spatial} sense is still significant. \looseness=-1

\section{Conclusion}
\looseness=-1 While classifier choice is known to be idiosyncratic, no extant study has precisely quantified this. To do so, we measure the mutual information between classifiers and other linguistic quantities, and find that classifiers are highly mutually dependent on nouns, but are less mutually dependent on adjectives and noun synsets.  Furthermore, knowing which noun or adjective supersense a word comes from helps, often significantly, but still leaves much of the original entropy in the classifier distribution unexplained, providing quantitative support for the notion that classifier choice is largely idiosyncratic. Although the amount of mutual dependence is highly variable across the semantic classes we investigate, we find that knowing a noun refers to a shape reduces uncertainty in classifier choice more than knowing it falls into any other semantic class, arguing for a role for information structure in Mandarin speakers' reliance on shape \citep{Kuo:09}. This result might have implications for second language pedagogy, adducing additional, tentative evidence in favor of collocational approaches to teaching classifiers \citep{zhang2013} that encourages memorizing classifiers and nouns together.

Investigating classifiers might also provide cognitive scientific insights into conceptual categorization \citep{lakoff1986}---often considered crucial for language use \citep{ungerer1996, taylor2002, croft2004}. Studies like this one opens up avenues for comparisons with other phenomena long argued to be idiosyncratic, such as grammatical gender, or declension class.

\section*{Acknowledgments}

This research benefited from generous financial support from a Bloomberg Data Science Fellowship to Hongyuan Mei and a Facebook Fellowship to Ryan Cotterell. We thank Meilin Zhan, Roger Levy, G\'eraldine Walther, Arya McCarthy, Ekaterina Vylomova, Sebastian Mielke, Katharina Kann, Jason Eisner, Jacob Eisenstein, and anonymous reviewers (NAACL 2019) for their comments. 

\bibliography{naaclhlt2019}
\bibliographystyle{acl_natbib}

\end{document}